\documentclass[preprint,12pt]{elsarticle}

\usepackage{amssymb}
\usepackage{enumitem}
\usepackage{framed,color}
\usepackage[algo2e,ruled,vlined,lined,linesnumbered]{algorithm2e}
\usepackage[algo2e]{algorithm2e} 
\usepackage{algorithm} 
\usepackage{algorithmic}   
\usepackage{multicol}
\usepackage{multirow}
\usepackage{setspace}
\usepackage[table]{xcolor}
\definecolor{shadecolor}{rgb}{1,0.8,0.3}
\usepackage{enumitem}
\usepackage{bussproofs}

\pdfoutput=1

\newtheorem{definition}{Definition}[section]

\newtheorem{example}{Example}

\newfont{\msbm}{msbm10 scaled 1000}

\def \tuple#1{\langle #1 \rangle} 

\def\defemb#1#2{\expandafter\def\csname #1\endcsname
                              {\relax\ifmmode #2\else\hbox{$#2$}\fi}}

\defemb{cA}{{\cal A}}
\defemb{cB}{{\cal B}}
\defemb{cC}{{\cal C}}
\defemb{cD}{{\cal D}}
\defemb{cE}{{\cal E}}
\defemb{cF}{{\cal F}}
\defemb{cG}{{\cal G}}
\defemb{cH}{{\cal H}}
\defemb{cI}{{\cal I}}
\defemb{cJ}{{\cal J}}
\defemb{cK}{{\cal K}}
\defemb{cL}{{\cal L}}
\defemb{cM}{{\cal M}}

\defemb{cO}{{\cal O}}
\defemb{cP}{{\cal P}}
\defemb{cQ}{{\cal Q}}
\defemb{cR}{{\cal R}}
\defemb{cS}{{\cal S}}
\defemb{cT}{{\cal T}}
\defemb{cU}{{\cal U}}
\defemb{cV}{{\cal V}}
\defemb{cX}{{\cal X}}
\defemb{cW}{{\cal W}}

\long\def\comment#1{}

\journal{Draft}

\begin{document}

\begin{frontmatter}



\title{How linguistic descriptions of data can help to the teaching-learning process in higher education, case of study: artificial intelligence}

\author[a1]{Clemente Rubio-Manzano}
\ead{clrubio@ubiobio.cl}
\author[a1]{Tom\'as Lermanda Senocea\'in}

\address[a1]{Dep. of Information Systems, University of the B\'{\i}o-B\'{\i}o, Chile.}

\begin{abstract}
Artificial Intelligence is a central topic in the computer science curriculum. 
From the year 2011 a project-based learning methodology based on computer games has been designed and implemented into 
the intelligence artificial course at the University of the B\'io-B\'io. The 
project aims to develop software-controlled agents (bots) which are programmed by using heuristic algorithms
seen during the course. This methodology allows us to obtain good learning results, however several challenges 
have been founded during its implementation. 

In this paper we show how linguistic descriptions of data can help to provide students and 
teachers with technical and personalized feedback about the 
learned algorithms. Algorithm behavior profile and a new Turing test for computer games bots based on linguistic 
modelling of complex phenomena are also proposed in order to deal with such challenges.  

In order to show and explore the possibilities of this new technology, a web platform has been designed and implemented 
by one of authors and its incorporation in the process of assessment allows us to improve the teaching learning process.
\end{abstract}

\begin{keyword}
Computational Intelligence, Linguistic Descriptions of Data, Linguistic Modelling of Complex Phenomena, Computer Game Bots, Turing Test, Computer-assisted Assessment 
\end{keyword}

\end{frontmatter}

\section{Introduction}

Feedback is an indispensable component of an effective teaching and learning environment in education
\cite{HattieTimperley2007, NicolMacfarlaneDick2006,SarsarHarmon2017,VandeKleij2012}.
Additionally, its personalization offers possibilities to deliver feedback that is the most appropriate for the
user's expertise, cognitive abilities and to their current moods and attentiveness \cite{Hamilton2009}. However,
providing students with personalized and immediate feedback is a complex task and it is usually a
standardized process (every student receives the same feedback, e.g., knowledge of correct response) due
to the large number of students \cite{Hamilton2009}. In writing skill, for example, truly immediate feedback is
impractical \cite{Lavolette2015}.

A possible solution to solve this limitation is to employ computer-assisted assessment (CAA). CAA is a longstanding
problem that has attracted interest from the research community since the sixties and has not been
fully resolved yet \cite{Perez2005}. The main aim is to study how the computer can help in the evaluation
of students' learning process \cite{Marin2004}. The literature has exposed several advantages:

\begin{itemize}[noitemsep]
\item It provides educators with didactic advantages \cite{VandeKleij2012}.
\item It provides students with immediate information in a timely manner and it is particularly useful when
the number of students is high and resources are scarce \cite{Hamilton2009}.
\item It is a quick way of providing feedback and it reduces the teacher's workload \cite{Lavolette2015}
\item It can be personalized, hence allowing the process of assessment to be enhanced from both teacher's
and students' points of view
\end{itemize}

Automatic assessment methods can be grouped into five main categories: statistical, natural language
processing (NLP), information extraction (IE), clustering and integrated-approaches \cite{NoorbehbahaniKardan2011}. 
Several examples of successful applications can be found in the literature:

\begin{itemize}[noitemsep]
\item Automatic creation of summaries assessment for intelligent tutoring systems \cite{Perez2005};
\item Automatic generation of formative feedback in the university classroom for specific concept maps
	scaffold students' writing \cite{Lachner2017}
\item A framework to provide students with feedback on algebra homework in middle-school classrooms
	\cite{Fyfe2016};
\item Automatic test-based assessment of programming \cite{Douce2005};
\item Automatic assessment of free text answers using a modified BLEU algorithm \cite{NoorbehbahaniKardan2011}.
\item Feedback for serious computer games to provide learners with useful and immediate information
	about the player's performance \cite{Burgos2009}.
\end{itemize}

The use of CAA in an undergraduate course of artificial intelligence can be very beneficial when a project-based learning 
is employed as teaching-learning methodology. An important skill to be acquired by undergraduate students of artificial intelligence 
courses is to get a better understanding of the different kind of heuristic algorithms existents for implementing computer games bots. 
In such project, each student individually \footnote{For us, the projects must be individually developed because team-based learning 
could has certain limitations
when it is applied for acquiring programming skills, however this discussion is out of the scope of this paper} must 
design and implement a computer game by programming the artificial intelligence of the various 
agents (bots) acting in 
the virtual world. This kind of project can be seen as a real computer game-based learning \cite{Prensky2007}. 

Computer Game-based learning \footnote{Note that, we do not use computer games for learning, students design and 
implement a computer game, that is,
the computer game is the result, it is not used as a pedagogical resource}, which is a type of CCA tool, was selected as a 
learning strategy because video-games 
are now used as new and powerful platforms for teaching and learning. In fact, the development of video games is currently a 
very motivational topic for the computer science students. 

In this context, the classical assessment of a computer 
game-based learning 
project consists in checking if the bots developed by the students are correctly designed and implemented. This process has 
important flaws: 

\begin{enumerate}[noitemsep]
\item It is a time-wasting task, mainly due to the excessive time required by the teacher to check the project's 
functionality. This becomes a serious problem when the number of students is high and there is only one teacher.
\item It is a complex task mainly due to the difficulty of evaluating a lot of important details about the 
implementation, which are usually missed in an execution trace: quantity (memory occupied, iterations performed, data 
structure used, etc.) and quality (how the artificial intelligence agent is good at capturing coins in the virtual world: is 
it fast, brave, intelligent?).  Additionally, both of them -quantity and quality- are difficult to capture due to the nature of 
the computer algorithm: they are running very fast, the debugger generates a lot of information which is difficult to understand, also 
large amount of data generated in the execution of a program. 
\item Impossibility (or very difficult) of performing individual project-based learning. 
\end{enumerate}

In the literature, some works have been proposed to provide learners and/or teachers with a CAA tool based on linguistic descriptions
of data generated into the learning process:
Automatic Textual Reporting in Learning Analytics Dashboards \cite{Ramos-Soto2017}, Feedback reports for students based on several 
performance factors \cite{Gkatzia2013} and
Reports describing the learner's rating in a specific learning activity \cite{Sanchez2012}.

In \cite{RubioTrivino2016} linguistic descriptions were used for improving player experience in a computer game
called YADY (Your actions define you). There are remarkable differences with respect to the present work. 
While in \cite{RubioTrivino2016} the feedback aims to improve player experience, now the feedback aims to 
support the teaching-learning process. 

In this paper, we propose a methodology and a data-driven software in order to automatically generate personalized 
and technical feedback from the data generated during the heuristic algorithm execution. A combination of three computational 
techniques is proposed, 
namely: bot's behaviour analysis, computational perception networks and natural language 
generation based on templates. The idea is that each student can gets immediate, technical and personalized feedback about their 
faults committed during the development of the project and they can learn about the heuristic 
algorithms employed for programming computer games bots.

On the other hand, this approach is very beneficial to the teachers since allows them to:

\begin{enumerate}[noitemsep]
	\item Save time for evaluating others aspects of the projects what implies a better understanding of them.
	\item Enhance the classical process of assessment providing students with  personalized and technical feedback. 
	\item Support individual project-based learning in order to get a more closed tracing of the projects and 
	the opportunity of focusing on the weak skills of the students and its strengthening.
\end{enumerate}

In order to show and explore the possibilities of this new technology a web platform has been designed and implemented by one of the 
authors following the phases and steps indicated in the methodology 
and the software specification (see User's Manual in Appendix). Additionally, this portal has been incorporated into the teaching learning process, now each student can consult the feedback in any time he/she wants and 
compare different kind of algorithms for programming computer games bots, also he/she can 
establish his/her own plan work for learning. Additionally, behavior profiles 
for computer game bots and human players allow to compare the
quality of the algorithms designed by students by using an adaptation of the Turing test which will be presented at \cite{Hingston2009}.

The structure of the paper is as follows. Section 2 introduces several
general concepts regarding project-based learning in artificial intelligence and provides a very brief review
of the state of art on the different involved disciplines. Then, in section 3
a methodology for incorporating linguistic description of data is proposed and incorporated into the AI projects.
Section 4 details the software architecture for providing teacher and students with personalized and technical feedback. Afterwards,
section 6 explains the experimentation and evaluation carried out on the projects
of the student by employing an adaptation of the Turing test. Finally, section 7
provides some concluding remarks.

\section{Preliminary Concepts} \label{preliminary}

\subsection{Linguistic Descriptions of Data and Natural Language Generation}

Linguistic Description of Data (LDD) intends to automatically produce expressions 
that convey the most relevant information contained (and usually hidden) 
in the data. It uses a number of modelling techniques taken from the soft 
computing domain (fuzzy sets and relations, 
linguistic variables, etc.) that are able to adequately manage the inherent 
imprecision of the natural language in the generated texts \cite{Ramos-Soto2015b}. 
LDD models and techniques have been used in a number of fields of application 
for textual reporting in domains such as: Deforestation Analysis \cite{Conde-Clemente2017a}, 
Big Data \cite{Conde-Clemente2017b}, Advices for saving energy at home \cite{Conde-Clemente2017c}, Self-Tracking 
Physical Activity \cite{Sanchez-Valdes2015}, cosmology \cite{Sanchez-Valdes2013,ArguellesTrivino2013}, 
driving simulation environments \cite{Esciolaza2013}, air quality index textual forecasts 
\cite{Ramos-Soto2015a}, weather forecasts \cite{Ramos-Soto2013}. It is a 
subfield of Artificial Intelligence (AI) which allows us to produce language as 
output on the basis of data input. 

NLG models and techniques have been applied for textual reporting in various 
domains, such as meteorological data \cite{Golberg1994,Coch1998}, care data \cite{Portet2009}, 
project management \cite{WhiteCaldwell1998}, air quality \cite{BusemannHoracek1997} 

\subsection{Restricted Equivalent Functions}

A restricted equivalent function (REF) \cite{BBP06} is a function which allows to establish a similarity between the 
elements of a domain. A REF can be formally defined as follows: 

\begin{definition}
	A REF, $f$, is a mapping $[0,1]^2 \longrightarrow [0,1]$ which satisfies the following conditions:
	
	\begin{enumerate}
		\item $f(x,y)=f(y,x)$ for all $x,y \in [0,1]$
		\item $f(x,y)=1$ if and only if $x=1$
		\item $f(x,y)=0$ if and only if $x=1$ and $y=0$ or $x=0$ and $y=1$
		\item $f(x,y)=f(c(x),c(y))$ for all $x,y \in [0,1]$, c being a strong negation.
		\item For all x,y,z $\in [0,1]$, if $ x \leq y \leq z$, then $f(x,y) \geq f(x,z)$ and $f(y,z) \geq f(x,z)$ 
	\end{enumerate}  
\end{definition}

For example, $g(x,y)=1-\left| x - y \right|$ satisfies conditions (1)-(5) with $c(x)=1-x$ for all $x \in [0,1]$. A similarity measure based
on REFs between linguistic terms has been recently proposed 
in order to enhance the inference engine of Bousi Prolog \cite{Rubio2017}.

\subsection{Project-based learning in Artificial Intelligence} 
\label{project}

From the year 2011 a project-based learning methodology based on computer games is applied into 
the intelligence artificial course at the University of the B\'io-B\'io. This methodology aims to
provide students with a better understanding of the heuristic algorithms which can be employed in real world applications.
To this end, the project aims to develop a computer game in which the bot's ability should be like that of the human players, 
being the programming skills and the abilities for incorporating 
them in the computer game very important competencies to be achieved as well.	

In the year 2017 the project consisted in the development of a set of computer game bots which should be designed and implemented by 
using the Java programming language. 
A computer game bot aims to remain itself inside of a scenario based on cells during the most
time possible. The student must take into account that the bots can lose energy in each movement performed (1 point of 
energy each five seconds). Three
Bots opponent (also programmed by the students) will treat to stole its energy and a set of rewards will be distributed at 
the scenario which for providing bots with additional energy.

\begin{figure}[t]
	\centering
	\includegraphics[height=4.5cm]{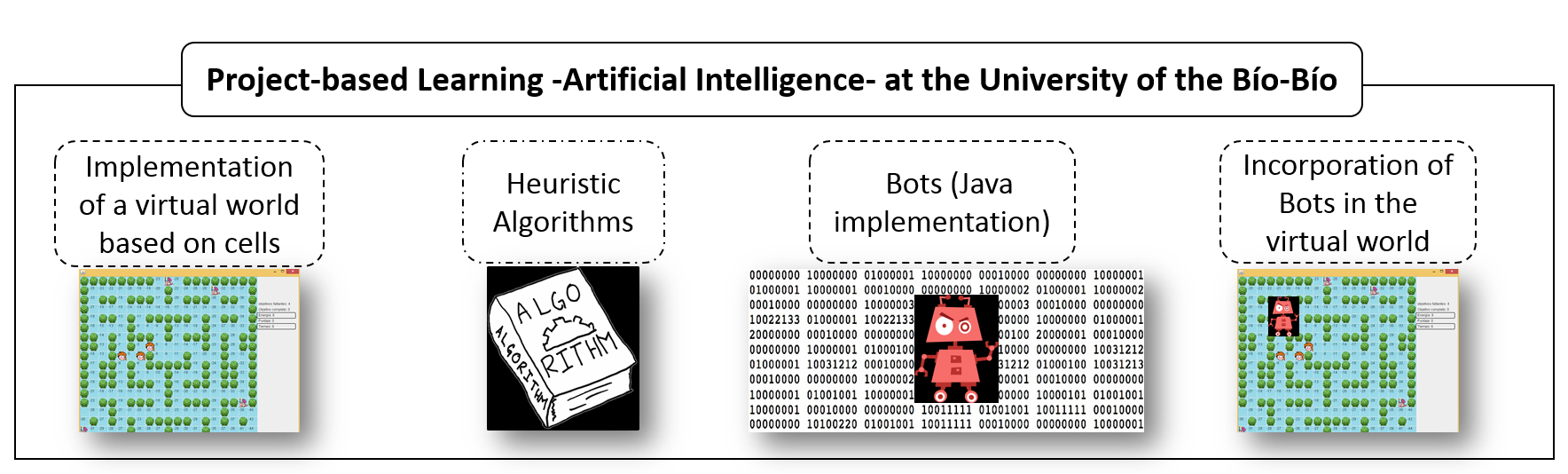}
	\caption{Project-based learning employed in the course of Artificial Intelligence at the University of the B\'io-B\'io}
	\label{fig-project}
\end{figure}

The first competence to be acquired for students in this project is the well-known algorithm thinking 
competence which is a pool of abilities for construction, analysis, specification and understanding of algorithms for 
solving given problem. Additionally, improving and adapting algorithms for given problems is an important ability 
to be acquired as well. In particular, the learning methodology employed in the artificial intelligence course at the University 
of the B\'io B\'io is as follows (see Figure~\ref{fig-project}):

\begin{enumerate}
	\item \textbf{Theoretical explanation} of the Heuristic algorithm is provided to students in order that they can get a better understanding 
	of them from a theoretical point of view. 
	\item \textbf{Implementation} of the algorithms must be performed by the students by using a particular programming language (Java is currently used) 
	in order that programming skills can be acquired by the students.
	\item \textbf{Understanding behavior of the algorithm} in a real-life context. The implementation performed by 
	the student is incorporated into the computer game. 
	\item \textbf{Evaluation of the Bots} created by the students is performed by checking if the bots is acting in a similar 
	way than human expert players. 
\end{enumerate}

We are going to pay attention on the third and the fourth items because an important question here is how a student of an 
artificial intelligence course can be sure that 
his/her designed and implemented bots is correctly working when it is incorporated in the computer game. An informal way to get it 
is by observing to the bot and to check that it is performed all
the functionalities. 

A limitation of this process is that some details about the design and the implementation could be lost due to the large amount
of data generated during the execution of the algorithm. This fact makes difficult to get an optimal assessment of the projects turning
the pedagogical monitoring of the students into a complex task. 

In order to address this flaw the concept of ''algorithm behavior profile'' is proposed. This profile is formed by the 
linguistic descriptions automatically generated by analyzing the data generated during its execution, 
then ''computer game bot behavior'' is defined as the behavior of a bot which has been implemented by using heuristic algorithms. 

Then, as LDD allows us to automatically generate a 
human behavior player profile \cite{RubioTrivino2016},  algorithm behavior profile can be obtained. 
By using this idea, the students can check 
if the designed computer game bot has a similar behavior than the human player one. In order to formally
define this comparison a Turing test based on LDD and REF for comparing profiles is proposed and 
explained in detail in section~\ref{turing}. 

\section{Methodology for incorporating linguistic description of data in AI projects} 

\begin{figure}[t]
	\centering
	\includegraphics[height=5.5cm]{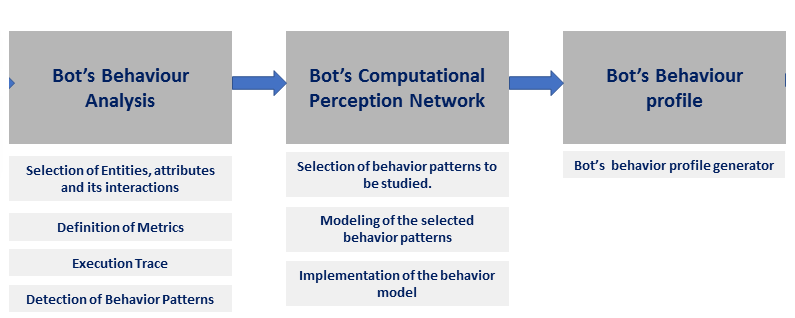}
	\caption{General methodology for automatically generated personalized feedback by using linguistic reports}
	\label{fig-architecture}
\end{figure}

This methodology aims is to provide us with a guide for creating a data-to-text system which transforms data 
in linguistic descriptions about the behavior of the computer bots implemented by using heuristic algorithms. First, data 
generated by the movements of the computer games bots will be grouped in metrics. The values 
captured for each metric will be clustered in linguistic terms. Next, a set of if-then rules will be
generated by aggregating these linguistic terms. The final result is a bot's behavior profile report automatically 
generated. 

The proposed methodology is formed by three phases: Bot's Behaviour Analysis, Linguistic Descriptions of Data and 
Evaluation (see Figure~\ref{fig-architecture}).

\subsection{Phase 1. Bots's Behaviour Analysis }

\begin{figure}[t]
	\centering
	\includegraphics[height=6.0cm]{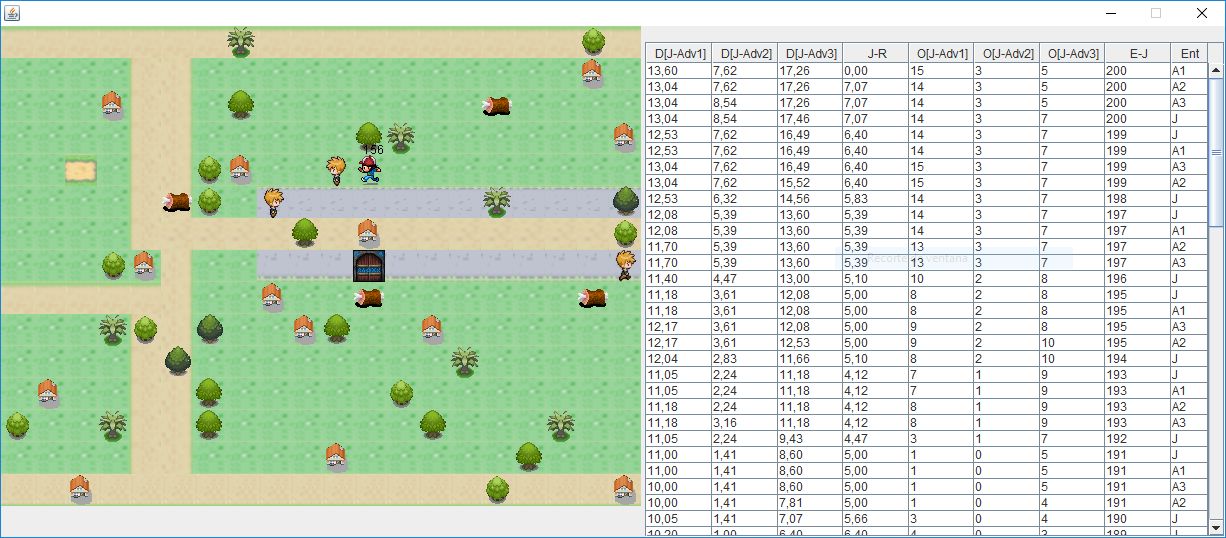}
	\caption{Trace of execution created from the values captured during the execution of the algorithms}
	\label{fig-trace}
\end{figure}

The aim of this phase is to analyze actions performed by the intelligent agents in order to get a set of 
behavior patterns.  In order to do that, the following steps can be followed:

\begin{enumerate}
\item \textbf{Selection of Entities, attributes and interactions} the main features of the entities which are acting in the virtual 
world must be selected in order to capture useful information about which entities must be taken into account in the process 
of assessment. To this end, entities, interactions and attributes must be identified and well-established.  

In our case, four entities have been identified: agent, opponents, rewards and obstacles. The agent has 
three important attributes: position x and y, energy and time employed to capture each reward. The rest of the entities and 
its attributes are shown in the Figure~\ref{fig-tabla-entidades}.

\begin{figure}[t]
	\centering
	{\footnotesize
	\begin{tabular}{|p{2.5cm}|p{2.5cm}|p{2.5cm}||p{5.0cm}|}
		\hline
		Entity & {\sf Attriibute} & {\sf Description} & {\sf Interactions and its effects} \\
		\hline
		Agent (1)   &  $Agent\_x$ & Position $x$  & Opponents (stole energy), Obstacles (protection), Rewards(gain energy) \\\hline
		         &  $Agent\_y$ & Position $y$ &  \\\hline
		         &  $Agent\_e$ & Energy $e$   & \\\hline
		Opponent(3) &  $Opponent\_x$ & Position $x$  & Obstacles,Rewards   \\\hline
		         &  $Opponent\_y$ & Position $y$  &  \\\hline
		Reward(4)   &  $Reward\_x$ & Position $x$  & Obstacles,Rewards   \\\hline
		         &  $Reward\_y$ & Position $y$  &  \\\hline
		Obstacle (M) &  $Obstacle\_x$ & Position $x$ & Obstacles,Rewards   \\\hline
		             &  $Obstacle\_y$ & Position $y$ & Obstacles,Rewards   \\\hline
		\hline
	\end{tabular}
}
	\caption{Entities, attributes and interactions. The number (i) is indicating the number of entities in the scenario. An agent, three opponents, four rewards and $M$ obstacles are the entities of the computer game}
	\label{fig-tabla-entidades}
\end{figure}

\item  \textbf{Definition of Metrics (Quantity and Quality).} From the entities selected in the previous step, a set 
of metrics can be defined in order to analyze its behavior. We are going to split between quantity metrics and 
quality ones. Quantity provides us with information about the performance of the algorithms (memory occupied and
iterations performed). On the other hand, quality provides us with information about the behavior of the heuristic algorithms, that is, 
how good
an algorithm is for implementing a computer game bot which should act like a human player. These metrics are defined from the entities and 
attributes identified in the previous item. The Figure~\ref{fig-tabla-metricas} shows these metrics and the corresponding descriptions.

\begin{figure}[t]
	\centering
	{\footnotesize
	\begin{tabular}{|p{2.5cm}|p{10.0cm}|}
		\hline
		{\sf Metric} & {\sf Description} \\
		\hline
		$Protection$ &  Number of obstacles between the agent and the $opponent_i$, a rectangular area is created from the position of the agent and the $opponent_i$, respectively\\\hline
		$Distance$ & Distance between two entities $E_1$ and $E_2$
		$d(E_1,E_2)=\sqrt{x-x')^2 + (x-x')^2}$, being $(x,y)$ the position of $E_1$ and $(x',y')$ the position of $E_2$ \\\hline
		
		$Energy$ & Energy of the player in an instant of time during the play session \\\hline
		
        $Time$ & Time registered from the start of the play session to the end of it \\\hline
		
		$Reward$ & True or false if a reward was captured at this instant of time \\\hline
		
		$Iterations$ & Number of iterations performed for the execution of the heuristic algorithm (It is executed in each move) \\\hline
		
		$Memory$ & Amount of memory required for the execution of the heuristic algorithm (It is executed in each move) \\\hline
		\hline
	\end{tabular}
}
	\caption{Metrics defined for analyzing the behavior of the heuristic algorithms}
	\label{fig-tabla-metricas}
\end{figure}

\item \textbf{Definition of a Computational Procedure to capture numerical data.} We are going to design and 
implement a procedure for capturing the data and it assigns values captured during a play session to the metrics 
defined in the previous task. A simple algorithm for capturing data can be performed in order to put them into a data structure 
which allows us to handle data in an efficient way.  

In our case, traces of execution (Figure~\ref{fig-trace}) have been employed as computational procedure for capturing and storing data. 
Tracing 
recording, or tracing is a commonly used technique useful in debugging and performance analysis. Concretely, trace 
recording implies detection and storage of relevant events during run-time, for later off-line analysis. We use a 
trace recording which stores the metrics defined in the previous item. The result is stored in a text file contained values for each 
metric defined in Figure~\ref{fig-tabla-metricas}. A set of execution traces can be found at the portal web (see User's Manual in Appendix).

\item \textbf{Detection of Behaviour Patterns.} Basic behaviour patterns can be established on the input data captured. An agent 
behavior pattern is associated with actions, that is, when a set of actions $\tuple{act_1, act_2,\ldots, act_n}$ are produced 
then a set of effects $\tuple{effect_1, effect_2,\ldots, effect_n}$ are trigged,e.g., when the opponent is close to 
the player then the player goes far away from he/she, so player and opponent are related and could provide us with 
some interesting behaviour pattern. Note that, patterns are related with the metrics defined in the previous item and 
they should created from them.
\end{enumerate}

\begin{figure}[t]
	\centering
	{\footnotesize
	\begin{tabular}{|p{3.0cm}|p{7.5cm}|p{2.5cm}|}
		\hline
		{\sf Behavior Pattern} & {\sf Description of the actions} & {\sf Metrics related} \\
		\hline
		Attitude & How the agent acts with respect to the reward, the distance between
		opponent and reward must be evaluated. & Distance   \\\hline
		Situation & How the agent acts with respect to the opponent, the energy and the protection must be evaluated  & Protection and Distance\\\hline
		Kind of move & Which is the result of a movement, distance between the agent and the reward and opponents must be evaluated & Distance and Energy \\\hline
		Performance & Which is the performance of each movement, time and memory must be
		measured & Memory and Iterations \\\hline
		\hline
	\end{tabular}
}
	\caption{Behavior Patterns created from the metrics defined in Figure~\ref{fig-tabla-metricas}}
	\label{fig-tabla}
\end{figure}

This phase provides us with a complete set of behavior patterns from actions performed by entities in the virtual world (see
Figure~\ref{fig-tabla-metricas}). It has 
been designed and implemented by using a computational perception network (see
Figure~\ref{report_generation}). 

\subsection{Phase 2. Linguistic Descriptions of Data in a 2D virtual world for automatically generating behavior profiles}

The aim of this phase is to establish a cognitive model from the previously behavior patterns identified and to generate 
linguistic descriptions about it. In order to do that the following step can be followed:

\begin{enumerate}[noitemsep]
	\item \textbf{Selection of behavior patterns to be studied.}  The behavior 
	patterns defined in the previous module are analyzed and selected according with our particular interest in them, that is, 
	which behavior patterns are important in order 
	to create the ''algorithm behavior profile''. For example, a particular sequence 
	of movement is not relevant for us, but the reason for performing it, it is really important. 
	
	\item \textbf{Modeling of the selected behavior patterns.} A cognitive model for treating 
	the patterns computationally is established. Taxonomies, ontologies, linguistic terms, if-then rules or a combination of them 
	could be used in this item without giving details about the implementation.
	
    \item  \textbf{Implementation of the behavior model.} The aim of this task is to  implement a computational solution for 
    representing the behavior model of our problem. Details about the implementation should be given. For example an "Ad-hoc" implementation could be employed or a package for automatically generating  linguistic descriptions could be used \cite{Conde-Clemente2017d}. It will depends on the kind of application to be developed. In our case, the PHP programming language has been employed because our ojective was the developement of a web platform and, in this context, PHP is a good alternative.  
	\item \textbf{Linguistic Descriptions Generator.} A linguistic description 
	generator is designed and implemented providing us with textual messages from the execution of the computational perception network. In our case,  linguistic summaries of data based on  fuzzy quantifiers have been employed which allows us to summarize the data generated in the process.
\end{enumerate}

This phase provide us with the behavior profile which is a graphical and textual describing the most relevant information about 
the behavior of the computer game bot during the execution of the algorithms used for its implementation.

\subsection{Evaluation: A Turing Test for Computer Game Bots based on LLD and REFs} \label{turing}

The analysis of algorithms for computer games is quite different that 
in others IA topics, while in the classical approach the aim is to simulate near-optimal 
intelligent behaviour, in computer games the aim is to provide interesting opponents for human players, 
not optimal ones \cite{Soni2008}.

As we mentioned the key in the evaluation in our artificial intelligence course is to develop computer game bots 
whose ability must be similar to the ability shown by a human player. 

How it could be formally evaluated? In \cite{Hingston2009} was proposed a variation of the Turing Test, designed to 
test the abilities of the computer game bots to impersonate a human player. The Turing test for (computer game) bots is 
as follows: \textit{''Suppose we are playing an interactive video game with some entity. Could you tell, solely from the conduct of the game, whether the other entity was a human player or a bot? If not, then the bot is deemed to have passed the test''}

This kind of Turing test is adapted to our methodology by using LLD and REF. The aim of this new Turing test is 
to establish a formal and effective method of comparison between the human 
behavior profile generated for a human expert player and the behavior bots profiles generated by a computer 
game bot implemented by using some heuristic algorithm. Therefore, a heuristic algorithm is 
near-behavioral (for us it is an ``optimal'' algorithm) when its associated profile is similar to the human expert 
profile. This novel process of assessment for heuristic algorithms will 
be detailed in the section~\ref{experimentation}.

\section{A data-driven software architecture based on Linguistic Modelling of Complex Phenomena}  

The software architecture proposed is formed for four modules which implements the phases and steps 
described in the methodology previously detailed:

\begin{enumerate}[noitemsep]
 \item Tracing module 
 \item Computational Perception Network module 
 \item Behavior Profile Report Generation module
 \item Evaluation module
\end{enumerate}

\subsection{Tracing module}

The tracing module aims to implement the functionality explained in the phase 1 once 
data have been, selected, captured and stored. The output of this module is a file contained the needed data captured 
during the execution of the algorithms implemented in the project (see portal web for more detail).

\subsection{Computational Perception Network module}
A computational perception network allows to implement the functionality explained in the phase 2. We use here 
the concept of Declarative Computational Perception(DCP) network which is inspired by 
the definition of CP network proposed~\cite{TrivinoSugeno2013} and it allows to model the 
problem in a declarative way. A declarative CP can be recursively defined as follows:

\begin{itemize}[noitemsep]
	\item \textbf{Base case.} A CP=(A,($u_1,\ldots,u_k$)), being A=($a_1, \ldots, a_n$) a vector of linguistic expressions 
	that represents the whole linguistic domain of CP whose values are calculated by aggregating	each $u_i$ 
	to either one or several elements of $A$.	
	\item \textbf{Inductive case.} A CP=(A,($CP_1, \ldots, CP_k$)), being A=($a_1, \ldots, a_n$) a vector of 
	linguistic expressions whose values are calculated by aggregating each $CP_i$ to either one or several elements of $A$.
\end{itemize}

Note that, the base case is produced when a CP is defined in terms of a real numbers set which belongs 
to a particular domain; i.e., a 1CP.\par 

The recursive case is produced when a CP is defined in terms of linguistic terms from a set 
of CPs; i.e., 2CP. We say that a set of sub-CPs $\{CP_{1},\ldots,CP_{k} \}$ completely define a CP or 
that a CP can be defined in terms of a sets of sub-CPs $\{ CP_{1},\ldots,CP_{k} \}$ . 

The computational perception presented in \cite{RubioTrivino2016} is enhanced for the problem previously 
presented in section~\ref{project}. In this case, additional variables must be considered and hence the computational perceptions 
network  must be enhanced, rules and templates must be also updated for these new 
requirements. 

Currently, a computer game bot can stay in a safe, easy, dangerous or risky situation, it depends on three factors, its
protection (low,normal,high) with respect to the opponent, the distance (close,normal,far) to the opponent and the 
energy (low,normal,high) that the bot has in this moment. Four attitudes can be detected for a computer game bot: wise, brave, cautious and passive. This depend on 
two factors, the distance between the bot and the closest reward and the distance between the opponent and the 
closest reward. A computer game bot can perform four types of movements: 
good, bad, scare, kamikaze. This depend on three factors, the distance between player and the closest reward,
the distance between the bot and the opponent and the energy of the bot. The ability of a computer game 
bot depends on its attitude, kind of movement performed and the time. The skill of a computer game bot depends on its attitude, 
kind of movement performed and situations detected. The resources (time and space) required for the execution of the 
algorithm used for implementing the artificial intelligence of the bot.
More formally, the computational network can be declaratively defined as follows (see Figure~\ref{report_generation} and 
appendix for more details).

{\footnotesize
\begin{itemize}[noitemsep]
\item $CP_{Situation}$=((Safe,Easy,Dangerous,Risky), ($CP_{Protection}^{player,opponent}$, $CP_{Distance}^{player,opponent}$, $CP_{Energy}^{player}$)). 
\item $CP_{Attitude}$=((Wise,Brave,Cautious, Passive), ($CP_{Distance}^{player,R*}$, $CP_{Distance}^{opponent,R*}$)) 
\item $CP_{Movement}$=((Good, Bad, Scare, Kamikaze), ($CP_{Distance}^{player,R*}$, $CP_{Distance}^{player,opponent}$, $CP_{Enery}^{player}$))
\item $CP_{Ability}$=((Expert, Intermediate, Basic, Dummy), ($CP_{Attitude}$, $CP_{Movement}$, $CP_{Time}$)) 
\item $CP_{Skill}$=((Very\_Skilled, Skilled, Improvable, Very\_Improvable), ($CP_{Attitude}$, $CP_{Movement}$, $CP_{Situation}$)) 
\item $CP_{Resources}$=((Very\_Efficient,Inefficient,Very\_Inefficient),($CP_{Iterations}, CP_{Memory}$))
\end{itemize}
}

\subsection{Behavior Profile Report Generation module}

The system selects, among the available possibilities, the 
most suitable linguistic expressions in order to describe the input data. 

We use $\Sigma CPs=((a_1,w_1), \ldots, (a_n,w_n))$ in order to generate a summarization of vector of linguistic expressions 
that represents the whole linguistic domain. These kind of CP allows us to obtain the total number of times in
which a value ($a_1, \ldots, a_n$) occurred during the execution. 

\begin{figure}
	\centering
	\includegraphics[height=9.5cm]{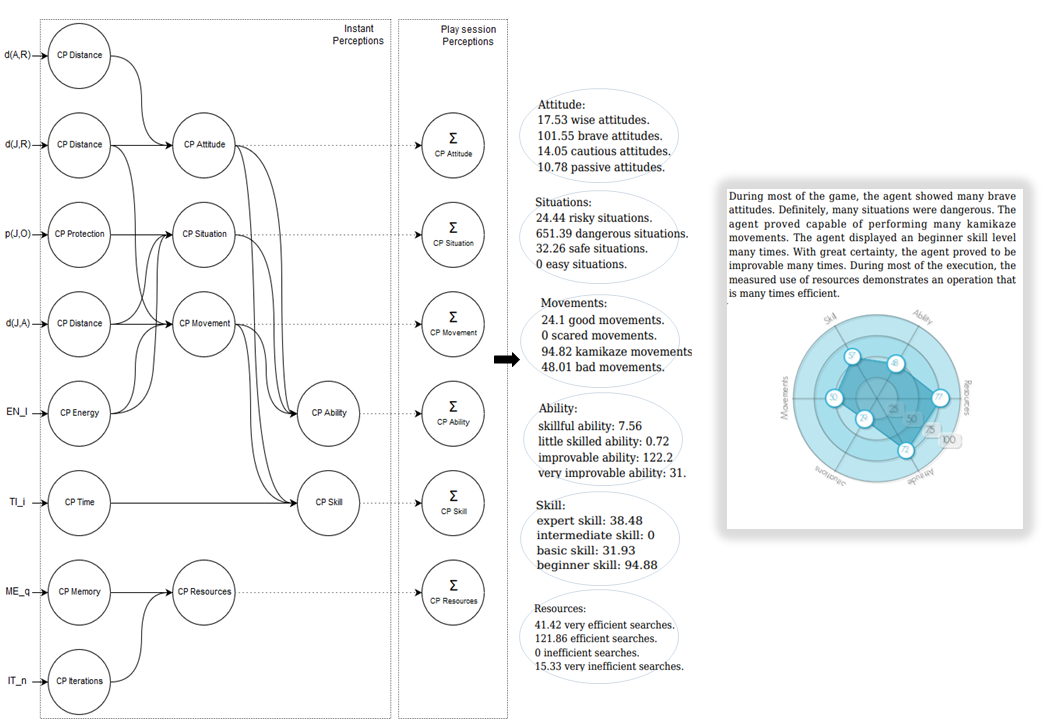}
	\caption{Automatic generation of a bot behavior profile: from trace of execution to linguistic descriptions}
	\label{report_generation}
\end{figure}

These kind of CP provides us with a set of variables, its 
associated value and a degree $\alpha$, which indicates the fuzzy average for a particular 
value. For example, a value for CP Situation could be $Safe$ with $0.8$ at an instant $i$ and $X=safe$ 
with $0.7$ at instant $i+1$, and so on. Therefore, at the end of the execution, we will have that $a_i$ (in 
the example ``safe'') has been given $N$ times with $N$ different degrees $\beta_1,\ldots,\beta_n$ (of course, some of these 
degrees could be equals). Thus, the final degree is calculated as follows: $\alpha_i=((\beta_1+ \ldots+\beta_n)/N)$.
For example, the following summaries can be obtained from different $\Sigma CP$ (see Figure~\ref{report_generation}).

The generation of the report is performed by using the set of $\Sigma CP$. For
each CP  a linguistic description is created in function of the pair $(a_i,w_i) \in \Sigma CP$. 
Percentages are calculated for each  $\Sigma CP$. The percentage
$p_i$ is then transformed in a linguistic term of quantity as follows: few is when $p_i \in [0,1/3]$; several when $p_i \in [1/3,2/3]$ or 
many when $p_i \in [2/3,1]$. Then, we are going to consider four cases:

\begin{enumerate}[noitemsep]
	\item There exists a pair $(a_i,p_i) \in \Sigma CP$ whose $p_i$ is greater than 66 percent 
	\item There exists a pair $(a_i,p_i) \in \Sigma CP$ whose $p_i$ is greater than 33 percent 
	\item There are two pair $(a_1,p_1),(a_2,p_2) \in \Sigma CP$ whose $p_i$ is greater than 33 percent 
	\item There not exists any pair $(a_i,p_i) \in \Sigma CP$  whose is greater 33 percent 
\end{enumerate} 

A complete example of behavior profile generation from data execution is detailed in Example~\ref{ejemplocompleto}.
\begin{example} \label{ejemplocompleto}
	Suppose a raw of the execution trace described in the Figure~\ref{fig-trace}.
	
	\begin{center}
		1, 13, 4, 12, 2, 12, 3.60, 3.16, 1.41, 17, 5.0, 2.0, 1.0, 15.26, 17.08, 13.0, 13.89, 15995,false. 42,
		924, J
	\end{center}
	
	The data are processed and grouped. In the Figure~\ref{fig-tabla-data-linguistic-terms} is shown the result. The second 
	column represents data captured and the third one terms linguistic created from 
	these data. Linguistic terms are implemented by using using fuzzy sets (trapezoidal functions), 
	membership degrees are also shown in the Figure~\ref{fig-tabla-data-linguistic-terms} with P=Position Player(1,13)
O1= Position Opponent1(3,16), O2=Position Opponent 2(4, 12) and O3=Position O3(2, 12).
	
	\begin{figure}[H]
		\centering
		{\footnotesize
			\begin{tabular}{|p{4.0cm}|p{1.0cm}|p{7.0cm}|}
				\hline
				{\sf Name} & {\sf Data} & {\sf Linguistic Term Generated} \\
				\hline
				Distance (P,o1) & 3.60 & Close (3.69, 0, 0, 4, 7) = 1 	\\\hline
				Distance (P,o2 & 3.16 & Close (3.16, 0, 0, 4, 7) = 1 	\\\hline
				Distance (P,o3) & 1.41 & Close (1.41, 0, 0, 4, 7) = 1 	\\\hline
				Energy(P) & 17 & High (17, 10, 13, 100, 100) = 1 	\\\hline
				Protection (P,o1) & 5.0 & High (5.0, 4, 6, 380, 380) = 0.5 	\\\hline
				Protection (P,o2) & 2.0 & Normal (2.0, 1, 3, 3, 5) = 0.5 	\\\hline
				Protection (P,o3) & 1.0 & Low (1.0, 0, 0, 0, 2) = 0.5 	\\\hline
				Distance (P,R*) & 15.26 & High (15.26, 13, 16, 38, 38) = 0.75 	\\\hline
				Distance (o1,R*) & 17.08 & High (17.08, 13, 16, 38, 38) = 1	 \\\hline
				Distance (o2,R*) & 13.0 & Normal (13.0, 6, 9, 11, 14) = 0.33 	\\\hline
				Distance (o3,R*) & 13.89 & High (13.89, 13, 16, 38, 38) = 0.29  	\\\hline
				Time & 15995 & Small (15995, 0, 0, 90000, 150000) = 1 	\\\hline
				Iterations at this movement & 42 & Normal (42, 18, 30, 42, 54) = 1  \\\hline
				Memory occupated(Bytes) & 924 & Low (924, 0, 0, 768, 1280) = 0.69  \\\hline
				\hline
			\end{tabular}
		}
		\caption{Data captured during a trace execution and the linguistic terms associated}
		\label{fig-tabla-data-linguistic-terms}
	\end{figure}

	Then each CP 
	is instantiated with the values of the linguistic terms and if-then fuzzy rules are computed by 
	using the average as t-norm of aggregation for computing  computational perceptions as follows (Obtained after processing and computing data in 
	Figure~\ref{fig-tabla-data-linguistic-terms}):
	{\small
	\begin{prooftree}
        \AxiomC{Distance(A,R*)=(High ,0.29)}
        \AxiomC{Distance(P,R)=(Normal, 0.33)}
        \AxiomC{Energy=(High,1)}
        \TrinaryInfC{Attitude=(Cautious,0.54)}
        \end{prooftree}
        
        \begin{prooftree}
        \AxiomC{Protection=(Low,0.5)}
        \AxiomC{Distance(P,O)=(Close,1)}
        \AxiomC{Energy = (High,1)}
        \TrinaryInfC{Situation=(Dangerous,0.83) }
        \end{prooftree}
	
	\begin{prooftree}
        \AxiomC{Distance(J,R*)=(Normal,0.33)}
        \AxiomC{Distance(P,O)=(Close,1)}
        \AxiomC{Energy=(High,1)}
        \TrinaryInfC{Movement=(Bad,0.91)}
        \end{prooftree}

        \begin{prooftree}
        \AxiomC{Attitude=(Cautious,0.54)}
        \AxiomC{Situation=(Dangerous,0.83)}
        \AxiomC{Movement=(Bad,0.91)}
        \TrinaryInfC{Ability=(Dummy,0.82)}
        \end{prooftree}

        \begin{prooftree}
        \AxiomC{Attitude= (Cautious,0.54)}
        \AxiomC{Movement=(Bad,0.91)}
        \AxiomC{Time=(Small,1)}
        \TrinaryInfC{Skill=(Improvable,0.76) }
        \end{prooftree}
	
	\begin{prooftree}
        \AxiomC{Memory=(Low,0.69)}
        \AxiomC{Iteration=(Normal,1)}
        \BinaryInfC{Resources=(Efficient,0.76)}
        \end{prooftree}
        }
Finally, the $\Sigma CP$ are computed:

\begin{itemize}[noitemsep]
\item $\Sigma CP_{Attitude}=\{$(wise,17.53), (brave,101.55), (cautious,14.05), (passive,10.78) $\}$
\item $\Sigma CP_{Situation}\{$(risky,24.44), (dangerous,651.39), (safe,32.26), (easy,0) $\}$
\item $\Sigma CP_{Movement}\{$(good,24.21), (scared,0), (kamikaze,94.82), (bad,48.01) $\}$
\item $\Sigma CP_{Ability}\{$(skillful,7.56), (little skilled,0.72), (improvable,122.2), (very improvable,31) $\}$
\item $\Sigma CP_{Skill}\{$(expert,38.48), (intermediate,0), (basic,31.93), (dummy,94.88) $\}$
\item $\Sigma CP_{Resources}\{$(very efficient,41.42), (efficient,121.86), (inefficient,0), (very inefficient,15.33) $\}$
\end{itemize}

From these $\Sigma CP$ and using the case established in a template, the instantiation is produced.
An example of template and its instantiation is showed in the Figure~\ref{report_generation}. The rest of the sentences 
for each CP are detailed in the appendix and a complete example is 
detailed in Figure~\ref{report_generation}

\begin{figure}[H]    
	{\footnotesize
		\begin{tabular}{|p{7.0cm}|p{7.0cm}|}	
			\hline			
			{\sf Template} & {\sf Instantiation after Bots play session}  \\\hline
			The bot showed $d_{Attitude}$ $a_{Attitude}$ attitudes. 
			Definitely, $d_{Situation}$ $a_{Situation}$ were safe. 
			The bot proved capable of performing $degree$ $value$ movements. 
			The bot displayed an $value$ skill level $degree$ times. 
			The bot proved to be $value$ $degree$ times. 
			During most of the execution, the measured use of resources demonstrates an operation that is $degree$ times $value$.  &  
			The bot showed \textbf{several brave} attitudes. 
			Definitely, \textbf{many} situations were \textbf{safe}. 
			The bot proved to be capable of performing \textbf{several good} movements. 
			The bot displayed an \textbf{expert} skill level \textbf{several} times. 
			The agent proved to be \textbf{skillful} \textbf{several} times. 
			During most of the execution, the measured use of resources demonstrates 
			an operation that is \textbf{many} times very \textbf{efficient} \\\hline
		\end{tabular}
	}
\end{figure}	
\end{example}

\section{Experimentation and Evaluation} \label{experimentation}

\begin{figure}
	\centering
	\includegraphics[height=6.0cm]{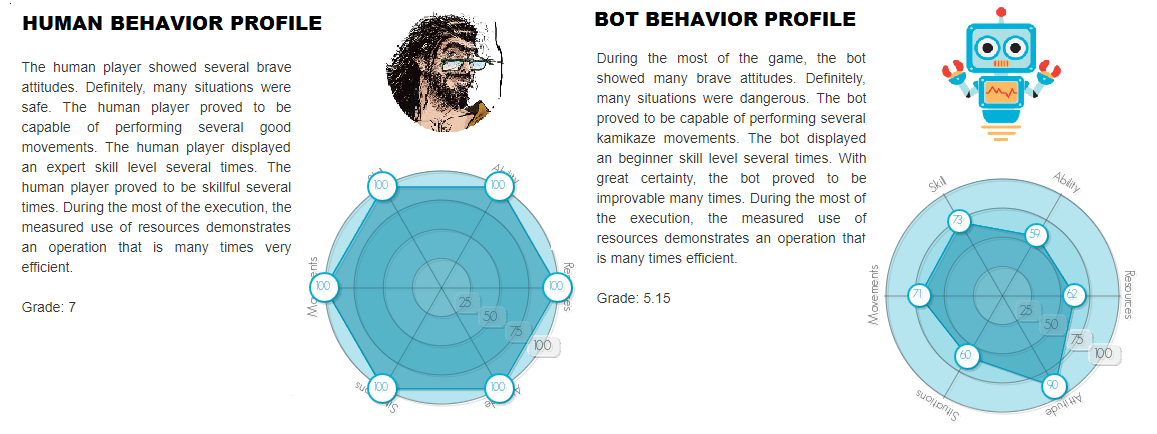}
	\caption{Similarity between hehavior profiles reports: human player versus bots}
	\label{fig-report}
\end{figure}

As we mentioned one the most important objectives in AI is to create an agent that simulates the human ability.
The bots behavior profile is compared with the human expert player profile (see Figure~\ref{fig-report}) by 
using a similarity measure based on REFs. The human player profile was created after that an expert human player 
played to the computer game with the following result:  

\begin{itemize} [noitemsep]
\item Attitude is mainly brave during the most part of the time.
\item Situation is mainly safe during the most part of the time.
\item Movements were mainly good during the most part of the time.
\item The player is expert.
\item The player is skilled.
\item The use of computational resource is efficient in time and space
\end{itemize}

The final grade (from 1 to 7) is computed by using the similarity between the human behavior profile and the bot one. The equation 
for calculating the final grade is as follows:
\begin{equation}
 FG=G_{Min} + S_{Attitude} + S_{Situation} +S_{Movement} + S_{Ability} + S_{Skill} + S_{Efficiency} 
\end{equation}

\begin{enumerate} [noitemsep]
	\item $G_{Min}$: 1 point (all the students have 1 point as a minimum score - it is mandatory at the University of the B\'io-B\'io)
	\item $S_{Attitude} = S_{REF}(\Sigma CP_{Attitude}^{Human},\Sigma CP_{Attitude}^{Bot})$ is the
	similarity between human player and bot attitude.
	\item $S_{Situation}= S_{REF}(\Sigma CP_{Situation}^{Human},\Sigma CP_{Situation}^{Bot})$: is the
	similarity between human player and bot situation.
	\item $S_{Movement}= S_{REF}(\Sigma CP_{Movement}^{Human},\Sigma CP_{Movement}^{Bot})$: is the
	similarity between human player and bot movements.
	\item $S_{Ability}= S_{REF}(\Sigma CP_{Ability}^{Human},\Sigma CP_{Ability}^{Bot})$:is the
	similarity between human player and bot ability.
	\item $S_{Skill}= S_{REF}(\Sigma CP_{Skill}^{Human},\Sigma CP_{Skill}^{Bot})$: is the
	similarity between human player and bot skill.
	\item $S_{Efficiency}= S_{REF}(\Sigma CP_{Efficiency}^{Human},\Sigma CP_{Efficiency}^{Bot})$:is the
	similarity between human player and bot efficiency.
\end{enumerate}

where $S_{REF}$ is a similarity measure between computational perceptions. The following definition 
formalizes this measure.

\begin{definition}
	Given two $\Sigma CP_i$, $\Sigma CP_j$ whose percentage linguistic vectors $\{(a_1,p_1)\ldots,(a_n,p_n)\}$ and 
	$\{(b_1,q_1)\ldots,(b_n,q_n)\}$ respectively. A similarity measure between $\Sigma CP_i$ and $\Sigma CP_j$ is defined 
	as:
	\begin{center}
	$S_{REF}(\Sigma CP_i, \Sigma CP_i)=\sum_{i=0}^{n} (REF(p_i,q_i)) / n$
	\end{center}
	being $REF(p_i,q_i)=1-\left| p_i-q_i\right|$
\end{definition}

\begin{example}
Let $CP_{Attitude}^{Human}, CP_{Attitude}^{Bot}$ be two summation computational perceptions for the human player and the computer 
game bot, respectively:

\begin{itemize} [noitemsep]
	\item  $\Sigma CP_{Attitude}^{Human}=\{$(wise,122.35), (brave,289), (cautious,87.59), (passive, 8.75) $\}$
	\item  $\Sigma CP_{Attitude}^{Bot}=\{$(wise,17.53), (brave,101.55), (cautious,14.05), (passive, 10.78) $\}$
\end{itemize}
	
Then, the percentages linguistic vectors are calculated for each $\Sigma CP$ by using their 
totals $Total_{\Sigma CP_{Attitude}^{Human}}(507.69)$ and $Total_{\Sigma CP_{Attitude}^{Bot}}(143.61)$, respectively:

\begin{itemize} [noitemsep]
	\item  $\Sigma CP_{Attitude}^{Human}=\{$(wise,0.240), (brave,0.569), (cautious,0.172), (passive,0.017) $\}$
	\item  $\Sigma CP_{Attitude}^{Bot}=\{$(wise,0.122), (brave,0.709), (cautious,0.097), (passive,0.075) $\}$
\end{itemize}

\end{example}

Now, the similarity $S_{REF}(\Sigma CP_{Attitude}^{Human},\Sigma CP_{Attitude}^{Bot})$ can be calculated:

\begin{itemize} [noitemsep]
	\item $REF(0.240,0.122)=1 - \left| 0.240-0.122\right|=0.882$
	\item $REF(0.569,0.172)=1 - \left| 0.569-0.172\right|=0.882$
	\item $REF(0.172,0.097)=1 - \left| 0.172-0.097\right|=0.925$
	\item $REF(0.017,0.075)=1 - \left| 0.017-0.075\right|=0.942$
\end{itemize}

Hence, $S_{REF}(\Sigma CP_{Attitude}^{Human},\Sigma CP_{Attitude}^{Bot})=\frac{3.402}{4}=0.838$. The rest
of the similarities is computed in a similar way. The final grade together with the linguistic 
reports generated for the human player and the bot designed by an anonymous student are shown in the Figure~\ref{fig-report}.

\section{Conclusions}

In this paper a novel and promising technology for automatically generating 
behavior profile reports and immediate feedback from the traces of execution of the heuristic algorithms has been presented. 

The concepts of the algorithm behavior profile has been proposed as 
a pedagogical tool for evaluating computer game bot quality. A Turing test based on the similarity between
bot behavior profile and human player has been defined and implemented.

These pedagogical resources provide teachers with an useful tool for getting information about 
the quality of the heuristic algorithm designed by the students what allows to improving the teaching and 
the learning process. The project created by the students can be evaluated in any time from two point of view: quantity 
(performance of the algorithm -space and time-), quality (kind of situations, movements, attitudes, abilities, skills).

As future work we would like to incorporate our technology in other high educational disciplines 
in order to obtain personalized feedback. 

\section*{Acknowledgments}

\footnotesize{
This work has been done in collaboration with the research group SOMOS (SOftware-MOdelling-Science) funded by the
the University of the B\'io-B\'io.
}

\appendix

\section{Definition of the Computational Perceptions}

\subsection{CP Situation}

{\footnotesize
This CP is defined as follows: $CP_{Situation}$=((Safe,Easy,Dangerous,Risky), 
($CP_{Protection}^{player,opponent}$, $CP_{Distance}^{player,opponent}$, $CP_{Energy}^{player}$)) 
with $CP_{Protection}^{player,opponent}$= ((low, intermediate,high),(0,1,$\ldots$,380))), with low(0,0,0,2),
intermediate(1,3,3,5) and high(4,6,380,380); $CP_{Distance}^{player,opponent}$=((close, normal, far),
(0,1,$\ldots$,38)), with close(0,0.4,7), normal(6,9,11,14) and dar(13,16,38,38)
}

\begin{figure}[H]    
{\footnotesize
	\begin{tabular}{|p{6.0cm}|p{8.0cm}|}	
		\hline	
		{\sf Consecuent (Situation)} & {\sf Antecedent (Protecction,Distance,Energy) } \\\hline		
		 Risky &  Intermediate, Close, Normal \\\hline
		Dangerous & Low, Close, Normal      \\\hline
		 Safe& Intermediate, Normal, Normal \\\hline
		 Easy& Low, Normal, Normal \\\hline
		 Dangerous & Low, Normal, Low      \\\hline
		 Dangerous & Normal, Close, Low      \\\hline
		 Dangerous & Normal, Normal, Low      \\\hline
	\end{tabular}
	}
\end{figure}	

\begin{figure}[H]    
{\footnotesize
	\begin{tabular}{|p{4.0cm}|p{10.0cm}|}	
		\hline			
		{\sf Cases for CP Situation} & {\sf Sentence }\\\hline
		1 &  Definitely, $degree$ situations were $value$\\\hline
		2 &  $degree$ situations were $value$ \\\hline
		3 &  $degree$ situations were $value_1$, although $degree$ situation also were $value_2$ \\\hline
		4 &  Diverse situations were detected during the most part of the play session \\\hline 
	\end{tabular}
	}
\end{figure}		

\subsection{CP Attitude}

{\footnotesize
This CP is defined as follows: $CP_{Attitude}$=((Wise,Brave,Cautious, Passive), ($CP_{Distance}$, $CP_{Distance}$)) 
and $CP_{Distance}^{Oponnet,R*}$= ((close, normal, far),(0,1,$\ldots$, size(scenario))) with close(0,0.4,7),  normal(6,9,11,14) and far(13,16,38,38), being R* the closest reward to the 
agent; $CP_{Distance}^{player,R*}$= ((close, normal, far),(0,1,$\ldots$, size(scenario))) with close(0,0.4,7), normal(6,9,11,14) and far(13,16,38,38), being R* the closest reward to the agent
}
\begin{figure}[H]    
{\footnotesize
	\begin{tabular}{|p{6.0cm}|p{8.0cm}|}	
		\hline	
		{\sf Consecuent(Attitude)} & {\sf Antecedent (Distance,Distance) } \\\hline		
		Wise &  Close, Normal \\\hline
		Brave& Close , Close    \\\hline
		Cautious & Normal, Close \\\hline
		Passive  &  Normal, Normal \\\hline
	\end{tabular}
	}
\end{figure}	

\begin{figure}[H]  
{\footnotesize
	\begin{tabular}{|p{4.0cm}|p{10.0cm}|}	
		\hline			
		{\sf Cases for CP Attitude} & {\sf Sentence }\\\hline
		1 &  During the most part of the play session, the bot showed $degree$ attitudes $value$ \\\hline
		2 &  The bot showed $degree$ attitudes $value$ \\\hline
		3 &  The bot showed $degree$ attitudes $value_1$, but also it showed $degree$ attitudes $value_2$ \\\hline
		4 &  The bot does not show a particular attitute during the play session'  \\\hline 
	\end{tabular}
	}
\end{figure}		

\subsection{CP Movement}
{\footnotesize
This CP is defined as follows: $CP_{Movement}$=((Good, Bad, Scare, Kamikaze), ($CP_{Distance}^{player,R*}$, $CP_{Distance}^{player,opponent}$, $CP_{Enery}^{player}$)), with $CP_{Distance}^{player,R*}$= ((close, normal, far),(0,1,$\ldots$, size(scenario)))  with close(0,0.4,7), normal(6,9,11,14) and far(13,16,38,38), being R* the closest reward to the agent; $CP_{Distance}^{player,opponent}$= ((close, normal, far),(0,1,$\ldots$, size(scenario))) with close(0,0.4,7), normal(6,9,11,14) and far(13,16,38,38), ;  $CP_{Enery}^{player}$=((low, normal, high),(0,1,$\ldots$, size(scenario))) 
}
\begin{figure}[H]   
{\footnotesize
	\begin{tabular}{|p{6.0cm}|p{8.0cm}|}	
		\hline	
		{\sf Consecuent(Movement)} & {\sf Antecedent(Distance,Distance,Energy) } \\\hline		
		Good  &  Close, Mormal, Normal \\\hline
		Good &  Close, Close, Low    \\\hline
		Scare  & Normal, Normal, Normal \\\hline
		Kamikaze & Close, Close, Normal  \\\hline
		Bad & Normal, Close, Normal \\\hline
	\end{tabular}
	}
\end{figure}

\begin{figure}[H]    
{\footnotesize
	\begin{tabular}{|p{4.0cm}|p{10.0cm}|}	
		\hline			
		{\sf Cases for CP Movement} & {\sf Sentence }\\\hline
		1 &  Certainly, $degree$ of the movements performed by the bot were $value$ \\\hline
		2 &  The bot proved to be capable of performing $degree$ movements $value$ \\\hline
		3 &  The bot proved to be capable of performing $degree$ attitudes $value_1$, but also performed $degree$ movements $value_2$ \\\hline
		4 &  The bot performs  indistinctly several movements during the play session \\\hline 
	\end{tabular}
	}
\end{figure}	

\subsection{CP Ability}

{\footnotesize
This CP is defined as follows: $CP_{Ability}$=((Expert, Intermediate, Basic, Dummy), ($CP_{Attitude}$,$CP_{Movement}$,$CP_{Time}$)) with $CP_{Time}=$((little,normal,large),(0,1,,$max\_time$)) with little, normal, and large
}

\begin{figure}[H]    
{\footnotesize
	\begin{tabular}{|p{6.0cm}|p{8.0cm}|}	
		\hline	
		{\sf Consecuent(Ability)} & {\sf Antecedent(Attitude,Movement,Time) } \\\hline		
		Expert & Wise, Good, Small \\\hline
		Intermediate   & Brave, Good, Normal \\\hline
		Basic& Passive, Bad, Much \\\hline
		Dummy& Passive, Scare, Much \\\hline
	\end{tabular}
	}
\end{figure}

\begin{figure}[H]    
{\footnotesize
	\begin{tabular}{|p{4.0cm}|p{10.0cm}|}	
		\hline			
		{\sf Cases for CP Ability} & {\sf Sentence }\\\hline
		1 &  Clearly, the bot displayed a/an $value$ player $degree$ times\\\hline
		2 &  The bot displayed a/an $value$  player $degree$ times \\\hline
		3 &  The displayed a/an $value$  player $degree_1$ times, however $degree_2$ times it acted
	as a/an $value_2$ \\\hline
		4 &  No kind of player has been identified \\\hline 
	\end{tabular}
	}
\end{figure}	

\subsection{CP Skill}
{\footnotesize
This CP is defined as follows: $CP_{Skill}$=((Very\_Skilled, Skilled, Improvable, Very\_Improvable), ($CP_{Attitude}$, $CP_{Movement}$, $CP_{Situation}$)) 
}
\begin{figure}[H]    
{\footnotesize
	\begin{tabular}{|p{6.0cm}|p{8.0cm}|}	
		\hline	
	   	{\sf Consecuent(Skill)} & {\sf Antecedent(Attitude,Movement,Situation) } \\\hline		
		Very\_Skilled & Wise, Good, Easy \\\hline
		Skilled  & Cautious, Good, Safe \\\hline
		Improvable & Brave, Bad, Dangerous \\\hline
		Improvable & Passive, Bad, Risky \\\hline
	\end{tabular}
	}
\end{figure}

\begin{figure}[H]    
{\footnotesize
	\begin{tabular}{|p{4.0cm}|p{10.0cm}|}	
		\hline			
		{\sf Cases for CP Skill} & {\sf Sentence }\\\hline
		1 &  Certainly, the bot proved to be $value$ $degree$ times \\\hline
		2 &  The agent proved to be $value$ $degree$ times \\\hline
		3 &  The agent proved to be $value_1$ $degree_1$ times, nevertheless $degree_2$ times proved
		to be $value_2$ \\\hline
		4 &  No kind of skill can be proved at the current play session \\\hline 
	\end{tabular}
	}
\end{figure}	

\section{User's Manual}

This section aims to explain in detail the use of the web platform for automatically generating human player 
and bot behaviour profiles from execution traces. A quick test of the application can be performed by 
downloading examples of traces at the following URL: 
\begin{center}
	\verb+http://youractionsdefineyou.com/assess/web/examples_traces+
\end{center}

First, the user must access to the URL: 
\begin{center}
	\verb+|http://www.youractionsdefineyou.com/assess+ 
\end{center}

\begin{figure}[H]
	\centering
	\includegraphics[height=8.0cm]{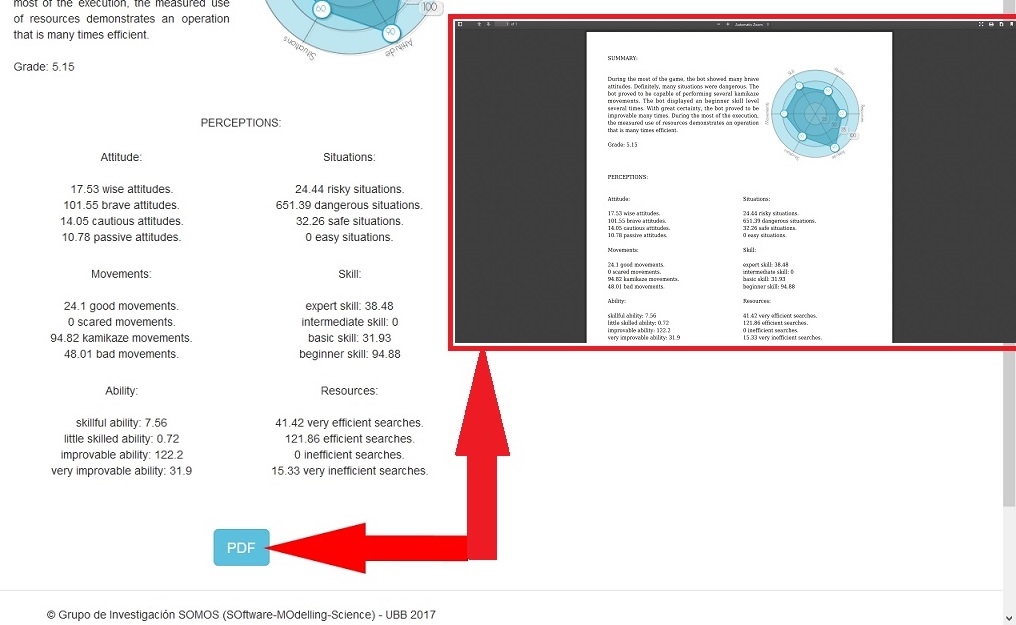}
	\caption{Linguistic Report obtained from a trace of execution of a human expert player}
	\label{fig-pdf}
\end{figure}

The main window shows two options:  log in and register. The register of a user consists in introducing email, 
user name, full name, RUT and password. A confirmation via email will be sent to the user if the registration was correct. 
The log in of a user consists in introducing the user name and the password.  
Second, behaviour profile report can be obtained by selecting and loading an execution trace file, then 
the behavior profile report is automatically generated. Additionally, the report can be exported in PDF (see Figure~\ref{fig-pdf}) 

\end{document}